\documentclass[sigconf,nonacm]{acmart} 
\usepackage{utfsym}
\usepackage{makecell}
\usepackage{pifont}
\usepackage{todonotes}
\usepackage{multirow}
\usepackage{listings}
\usepackage{adjustbox}
\usepackage{listings}
\usepackage{lipsum}
\usepackage{cleveref}
\usepackage{subcaption}
\usepackage{threeparttable}
\usepackage{fancyvrb}
\usepackage{enumitem}
\usepackage[most]{tcolorbox}
\tcbuselibrary{skins}

\definecolor{greyback}{gray}{0.96} 
\tcbset{insightbox/.style={
  enhanced,
  boxrule=0pt,
  colback=greyback,
  colframe=greyback,
  borderline west={2pt}{0pt}{black},
  left=8pt,
  right=8pt,
  top=6pt,
  bottom=6pt,
  before skip=10pt,
  after skip=10pt,
  fontupper=\normalsize,
}}

\usepackage{balance}
\usepackage{listings}
\usepackage[T1]{fontenc} 

\begin{document}
\title{Prompting Large Language Models for Training-Free Non-Intrusive Load Monitoring}

\author{Junyu Xue}
\authornote{Work done while interning at HKUST-GZ.}
\affiliation{
  \institution{Southern University of Science and Technology \& Peng Cheng Laboratory}
\country{}
}
\author{Xudong Wang}
\affiliation{
  \institution{The Chinese University of Hong Kong, Shenzhen}
  \country{}
  }
\author{Xiaoling He}
\affiliation{
  \institution{The Hong Kong University of Science and Technology (Guangzhou)}
  \country{}
}
\author{Shicheng Liu}
\affiliation{
  \institution{Tsinghua University}
  \country{}
}
\author{Yi Wang}
\affiliation{
  \institution{Southern University of Science and Technology \& Peng Cheng Laboratory}
  \country{}
}
\author{Guoming Tang}
\authornote{Correspondence to: Guoming Tang <guomingtang@hkust-gz.edu.cn>.}
\affiliation{
  \institution{The Hong Kong University of Science and Technology (Guangzhou)}
  \country{}
}

\begin{abstract}
Non-intrusive load monitoring (NILM) aims to disaggregate total electricity consumption into individual appliance usage, thus enabling more effective energy management. While deep learning has advanced NILM, it remains limited by its dependence on labeled data, restricted generalization, and lack of explainability. This paper introduces the first prompt-based NILM framework that leverages large language models (LLMs) with in-context learning. We design and evaluate prompt strategies that integrate appliance features, contextual information, and representative time-series examples through extensive case studies. Extensive experiments on the REDD and UK-DALE datasets show that LLMs guided solely by prompts deliver only basic NILM capabilities, with performance that lags behind traditional deep-learning models in complex scenarios. However, the experiments also demonstrate strong generalization across different houses and even regions by simply adapting the injected appliance features. It also provides clear, human-readable explanations for the inferred appliance states. Our findings define the capability boundaries of using prompt-only LLMs for NILM tasks. Their strengths in generalization and explainability present a promising new direction for the field.
\end{abstract}


\maketitle

\section{Introduction}\label{sec:introduction}

Non-intrusive load monitoring (NILM) aims to infer the energy consumption of individual appliances by analyzing aggregate electricity usage, thereby avoiding the need for sub-metering individual devices~\cite{hart1992nonintrusive}. 
By providing detailed appliance-level energy consumption data, the NILM technique empowers users with actionable insights for energy savings, enhances smart home automation, and facilitates energy demand management~\cite{zoha2012non}.
However, mathematically, decomposing the aggregated signal into individual appliance contributions is an ill-posed problem~\cite{hart1992nonintrusive}. This challenge requires powerful modeling approaches capable of learning intricate patterns from the data~\cite{kolter2011redd, makonin2015nonintrusive}.

\begin{figure}
    \centering
    \includegraphics[width=0.85\linewidth]{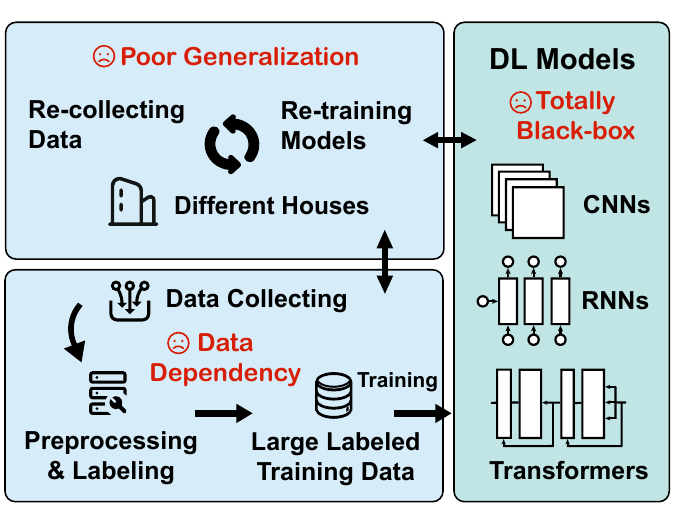}
    \caption{Limitations of traditional DL-based NILM methods.}
    \label{fig:limitations}
    \vspace{-0.55cm}
\end{figure}

Deep learning (DL)~\cite{lecun2015deep} has significantly advanced the field of NILM due to its ability to autonomously learn complex features and model intricate dependencies directly from raw aggregate power signals~\cite{kelly2015neural, du2016regression, zhang2018sequence, yue2020bert4nilm, yu2024auglpn}. 
Although DL-based NILM methods have demonstrated promising performance, further refinement and validation remain active research areas. As shown in Figure~\ref{fig:limitations}, the main limitations include data dependency, poor generalization, and lack of explainability. 
These limitations create a significant gap between the theoretical performance of NILM algorithms and their practical applicability in real-world scenarios~\cite{towardsreal-worldNILM}.

\begin{figure*}[htbp]
    \centering
    \includegraphics[width=0.95\linewidth]{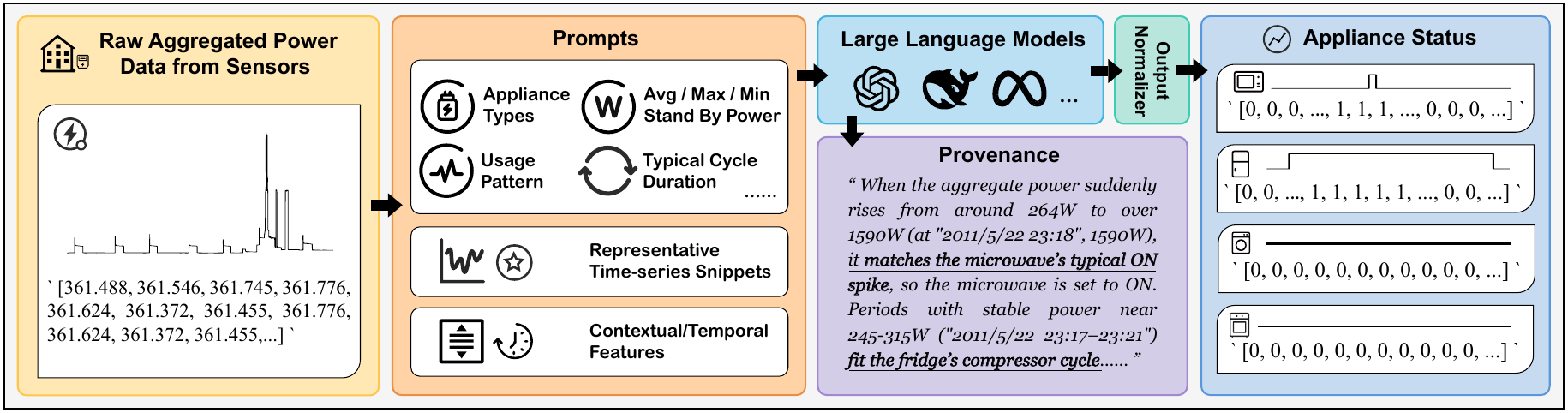}
    \caption{An overview of our proposed LLM4NILM framework.}
    \label{fig:overview}
    \vspace{-0.2cm}
\end{figure*}

Recently, large language models (LLMs) have demonstrated remarkable capabilities across a vast range of domains~\cite{zhao2023survey}, driven by advancements in natural language processing, reasoning, and particularly \textbf{in-context learning}~\cite{dong2024survey}.
LLMs can perform specific tasks based on instructions and knowledge provided within a prompt, and often achieve reasonable performance in zero-shot or few-shot settings without task-specific fine-tuning~\cite{wang2019survey,wang2020generalizing}.
These unique strengths motivate investigating their applicability to NILM, leading us to focus on the following three compelling and opening research questions (\textbf{RQs}): 

\begin{itemize}[leftmargin=*, labelsep=1em]
    \item \textbf{RQ1.} Can LLMs execute NILM effectively through prompt-based zero-shot learning, leveraging their inherent knowledge to circumvent the extensive data acquisition and model training phases in conventional DL approaches? 
    \item \textbf{RQ2.} Can LLMs achieve robust generalization in NILM across diverse environments (e.g., unseen houses with varying appliance types and numbers) via prompting?
    \item \textbf{RQ3.} Can LLMs enhance the explainability of the NILM process by generating verifiable explanations for their disaggregation outputs, thus mitigating the opacity characteristic of conventional DL methods?
\end{itemize}

In this paper, we explore the above three \textbf{RQs} by conducting a systematic experimental study that examines the effectiveness and potential of modern LLMs when applied to NILM tasks using prompt-based inference.
As Figure~\ref{fig:overview} shows, we design and implement a unified framework named \textbf{LLM4NILM}, which is centered on advanced prompt engineering.
Within this framework, we integrate various prior knowledge sources, such as appliance features, one-shot examples, timestamps, and contextual information from previous time windows. 


Then, we validate this framework with extensive case studies. We ablate each prompt component and knowledge type and study the sensitivity to window length and context length. These experiments confirm the design soundness and reveal the factors that most influence accuracy. 
In detail, our ablation studies show that incorporating contextual information from previous predictions provides the most significant performance boost. We also find that injecting domain knowledge is highly effective, with an appliance's power range being the most critical feature for its identification. 
Furthermore, our sensitivity analyses establish a crucial trade-off, showing that the input power sequence length (window size), the amount of historical state information (context length), and the LLMs' inherent capacity must be co-optimized to balance predictive accuracy with output stability.

Guided by these insights, we then apply the optimized prompts to the REDD \cite{kolter2011redd} and UK-DALE \cite{kelly2015uk} datasets to deliver informed answers to our \textbf{RQs}. 
Our evaluation on these large-scale datasets reveals a clear trade-off. The LLM-based approach, while not matching the accuracy of specialized deep learning models, demonstrates significant strengths in generalization and explainability. The model successfully generalizes across different houses and even regions by simply adapting the prompts without any retraining. It also enhances transparency by generating human-readable explanations for its predictions. These findings suggest the current value of LLMs is not in replacing trained models for pure accuracy. Their potential lies in complementing existing systems to build more adaptable and transparent hybrid solutions.

In summary, our main contributions are as follows:

\begin{itemize}[leftmargin=*, labelsep=1em]
\item To the best of our knowledge, we conduct the first systematic study applying prompt-driven, general-purpose LLMs to core NILM tasks, and propose a comprehensive framework for the methodical exploration and evaluation of this novel approach on standard benchmarks.
\item We demonstrate the effectiveness of our framework, showing that even with optimized prompt engineering, current LLMs struggle to achieve accuracy competitive with specialized deep learning models.
\item We establish that this LLM-based approach offers significant advantages in robust zero-shot generalization across diverse, unseen environments without task-specific fine-tuning, and enhances explainability by generating verifiable explanations for its disaggregation outputs.
\end{itemize}

\section{Related Work}\label{appendix:relatedwork}
\subsection{Non-intrusive Load Monitoring}
\paragraph{\textbf{Accuracy.}}
NILM is inaugurated by fingerprint methods that treat each switching event as a unique signature. NALM system models appliances as finite-state machines and matches step changes in real power and reactive power to those states \cite{hart1992nonintrusive}.
As larger data sets become available, researchers move to probabilistic graphical models, most notably the additive Factorial Hidden Markov Model (FHMM), which expresses aggregate demand as the sum of multiple independent appliance chains and applies convex variational inference for tractable disaggregation \cite{kolter2012approximate}. 
In recent years, DL approaches commonly incorporate artificial neural network components such as CNNs~\cite{lecun1998gradient}, RNNs~\cite{hochreiter1997long}, and Transformers~\cite{vaswani2017attention,yu2024auglpn}, which have consistently enabled state-of-the-art performance on public energy disaggregation datasets. But DL methods typically require \textbf{large amounts of labeled training data}, which is expensive and difficult to acquire at scale~\cite{rafiq2024review,kelly2015neural}.

\paragraph{\textbf{Explainability.}}
Although DL methods' high accuracy is promising, their opaque decision-making processes impede user trust. 
The majority of explainability research attempts to solve this by visualising saliency. 
One approach grafts a decision-tree surrogate onto a model to expose feature importance with PD/ICE plots~\cite{mollel2022using}. 
A complementary study develops a metric suite with faithfulness, robustness, and effective complexity to objectively compare heatmap-based explanations~\cite{batic2023toward}. 
Another work, targeting real-world applications, builds a high-frequency pipeline using SHAP that delivers sub-second attributions on edge devices~\cite{gerasimov2025toward}. 
Yet, all these methods produce technical graphics that demand expert analysis. 
LLMs point to a complementary solution. 
They can distil internal evidence into concise natural-language rationales and answer counterfactual queries, making NILM explanations simultaneously machine-verifiable and accessible to non-specialists.

\paragraph{\textbf{Generalization.}}
The ability to generalise to unseen appliances and new households is pivotal for real-world NILM applications. Labelled data remain scarce, and models trained on one site often fail elsewhere, prompting costly retraining or ad-hoc transfer learning pipelines ~\cite{li2023transfer,li2023adaptive,wang2022evsense}. Recent semi-supervised, unsupervised and pre-training strategies mine structure from un-labelled traces and already yield measurable cross-dataset gains, easing the data-scarcity bottleneck~\cite{qiu2024semi}. Federated optimisation extends the idea: by keeping data local while sharing model updates, horizontal and personalised variants~\cite{zhang2022fednilm,luo2025federated}. Yet these schemes still require intricate orchestration and can falter on truly novel device signatures. LLMs, with their in-context learning and compositional reasoning, hint at a complementary path by recognising new appliance patterns and household behaviours without gradient updates, thus opening a new frontier for NILM generalisation.

\begin{figure}
    \centering
    \includegraphics[width=0.75\linewidth]{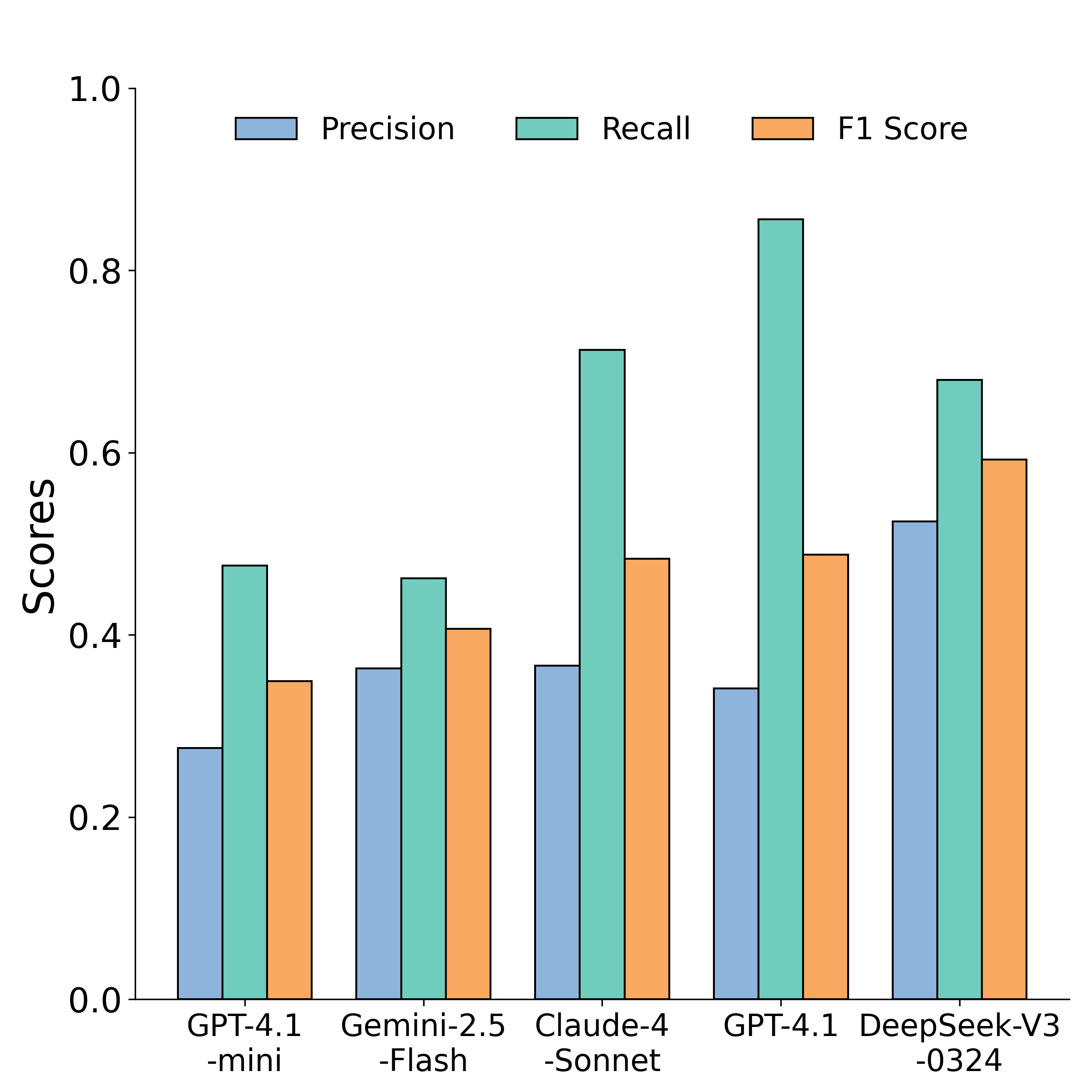}
    \caption{The NILM capability of mainstream general LLMs.}
    \label{fig:sota-models}
    \vspace{-0.2cm}
\end{figure}

\subsection{Large Language Models and Its Applications}
LLMs have surged in capability after scaling and instruction tuning, displaying in-context learning, reasoning, and emerging multimodal alignment~\cite{zhao2023survey}. These properties suggest a single foundation model can reason across language, vision, and sequential sensor data.

\paragraph{\textbf{Time-Series Tasks.}}
Recent work~\cite{xue2023promptcast, gruver2023large, zhang2024large} has applied LLMs to time-series problems, including zero-shot load forecasting and anomaly classification, by representing temporal windows as token sequences. Given that NILM fundamentally involves analyzing sequential power data, the demonstrated success of LLMs on time-series tasks underscores their significant promise for load disaggregation.

\paragraph{\textbf{Disaggregation Tasks.}}
LLMs have been established as effective decomposers, initially demonstrating the ability to break down complex textual problems into sequential reasoning steps~\cite{wei2022chain}. This paradigm now extends to other modalities, where LLMs orchestrate specialized models for tasks like audio source separation~\cite{wang2025unisep} and image object recognition~\cite{lai2024lisa}. The underlying principle of treating disaggregation as a sequential reasoning problem is directly applicable to NILM, which seeks to separate individual appliance consumption from an aggregate signal. However, compared to the audio and vision domains, the potential of LLMs as decomposers in NILM remains underexplored.

Overall, whether the reasoning and in-context learning capabilities that empower LLMs in other data modalities can translate into better generalization, data efficiency, and human-readable explanations for NILM is still an open problem. Our work serves as a pioneer in exploring how LLMs can be adapted for energy disaggregation.

\section{LLM4NILM: Framework and Prompt Design}\label{sec:promptingLLMsforNILM}
As mentioned in \textbf{RQs}, the core challenge is effectively conveying the domain knowledge required for NILM tasks via LLM through prompts. 
In this section, we first design the basic prompt strategy~(\Cref{subsec:foundationalPrompts}) to ensure that the LLM can perform NILM tasks. Then we explore more advanced prompt strategies, including incorporating timestamps~(\Cref{subsec:temporalPrompts}) and contextual information~(\Cref{subsec:contextPrompts}) for performance enhancement. The complete prompt can be found in Appendix~\ref{appendix:prompts}.

\subsection{Foundational Prompts}\label{subsec:foundationalPrompts}
For the design of foundational prompts for NILM, we draw upon established principles~\cite{zhao2023survey} and structure our prompts with several key components detailed below: a \textit{Base Prompt} defining the LLM's role and the core task, \textit{Knowledge Injection} providing domain-specific information, \textit{One-shot Examples} to illustrate the task, and specifications for the \textit{Input Data}.

\paragraph{\textbf{Base Prompt (Role \& Task).}}
An appropriate prompting strategy can stimulate the LLM's capabilities for specific tasks. Here, we inform the LLM of its role as an expert in the NILM domain.
The task description is crucial for guiding the LLM's understanding of the NILM task. We provide a clear and concise description of the task, including the goal of disaggregating total power consumption into individual appliance usage.
Additionally, we specify the required input and output formats and the expected time window length for the task.
It is worth noting that we set two tasks: state prediction and power prediction, which have different descriptions. The corresponding time window lengths may also differ for these tasks.

\paragraph{\textbf{Knowledge Injection.}}
We inject relevant knowledge about appliances to enhance the LLM's understanding of the NILM domain. The most basic requirement is to inform the LLM about the types of appliances involved in the task. For example, in the REDD dataset, we provide a list of appliances, involving refrigerator, washing machine, microwave and dishwasher.
In addition, we provide the \textbf{power consumption characteristics} of each appliance, which are extracted from historical data using statistical methods (e.g., standby power, ON power range, average ON duration, usage pattern, typical cycle duration, etc.).

\paragraph{\textbf{One-Shot Example.}}
We provide several examples to help the LLM understand how to transform input data into output results.
These examples include input data and corresponding output results. Intuitively, the more diverse the examples are regarding aggregated power consumption patterns, the better the model can learn from them.
However, due to the token limit of API calls, we can only provide a limited number of examples. The examples also help the LLMs understand the format and task requirements.

\paragraph{\textbf{Input Data.}}
The input is the original aggregated power consumption data. We format the input and output data in JSON format, consistent with the NILM system's approach~\cite{towardsreal-worldNILM}. 
In this paper, we enable the API's JSON mode and specify the corresponding format in the prompts.

The above four components constitute our foundational prompts for NILM. Through them, we can effectively guide the LLM in understanding the NILM task and provide necessary prior knowledge. 

\begin{table}[!t]
  \centering
  \setlength{\tabcolsep}{2pt}
  \begin{threeparttable}
    \caption{Performance and resource consumption of different prompts for NILM. $\uparrow$: higher is better.}
    \label{tab:prompts-results-nilm}
    \begin{tabular*}{\linewidth}{@{\extracolsep{\fill}} lcccc@{}}
      \toprule
      \textbf{Components}\tnote{1} &
      \textbf{Precision $\uparrow$} &
      \textbf{Recall $\uparrow$} &
      \textbf{F1-score $\uparrow$} &
      \textbf{\#Token} \\
      \midrule
      Base                  & 0.3475 & 0.5392 & 0.4226 &  747 \\
      Base+OE               & 0.3570 & 0.5398 & 0.4298 & 1004 \\
      Base+OE+KI            & 0.4021 & 0.5051 & 0.4477 & 1535 \\
      Base+OE+KI+TS         & 0.3896 & 0.6153 & 0.4771 & 1703 \\
      Base+OE+KI+CT         & \textbf{0.5246} & \textbf{0.6795} & \textbf{0.5921} & 2299 \\
      Base+OE+KI+TS+CT      & 0.4004 & 0.5960 & 0.4790 & 2480 \\
      \bottomrule
    \end{tabular*}
    \begin{tablenotes}
      \footnotesize
      \item[1] OE, KI, TS, and CT denote one-shot prompting example, knowledge injection, timestamps, and contextual information, respectively.
    \end{tablenotes}
  \end{threeparttable}
  \vspace{-0.3cm}
\end{table}

\subsection{Enhancing Prompts with Timestamps}\label{subsec:temporalPrompts}
In the design of foundational prompts, timestamps are not included by default, primarily because their inclusion incurs a substantial token overhead.
However, temporal features are valuable for NILM tasks, as the state and power consumption of appliances are often time-dependent. For example, dishwashers are rarely used in the early morning, and refrigerators typically exhibit periodic power consumption patterns.
Therefore, we explore the possibility of incorporating temporal features into the prompts. We include timestamp information (HH:mm:ss) in the prompts to help the LLM better understand appliance usage patterns.
Specifically, we modify the base prompt, input data, and one-shot example in the foundational prompts. We add external explanations for temporal features and include a timestamp field in each input data entry.

\subsection{Enhancing Prompts with Context}\label{subsec:contextPrompts}
When using LLMs for NILM, we typically rely on API calls for inference. However, simple API services do not provide contextual information.
As a result, the inferences for different time windows are independent and previous results cannot be leveraged to inform subsequent ones.
Yet, connections between time windows are inherent in NILM tasks. In traditional DL methods, researchers use filtering and other post-processing techniques to smooth the transitions between time windows.
Our approach incorporates the \textbf{inference results from the previous time window} into the prompt to establish this connection.

\section{Implementation}\label{sec:implementation}
Following the design principles for the LLM4NILM framework established in \Cref{sec:promptingLLMsforNILM}, this section outlines the implementation details of our evaluation. We describe the datasets, data processing methods (\Cref{subsec:datasetandpreprocessing}), evaluation metrics (\Cref{subsec:metrics}), the LLMs utilized, and our post-processing approach (\Cref{subsec:modelselect}).

\subsection{Dataset and Preprocessing}\label{subsec:datasetandpreprocessing}
We use the most widely used public benchmark datasets, REDD~\cite{kolter2011redd} and UK-DALE \cite{kelly2015uk} in this paper. REDD provides energy consumption recordings from 6 residential houses in the USA and includes both high and low frequency active power measurements for whole-house mains and individual appliances. UK-DALE contains detailed electricity measurements gathered from five typical residential households in the UK. It offers whole-house power data alongside readings for numerous individual appliances, captured at various sampling frequencies.

For preprocessing, we utilize the low-frequency active power data from selected houses in the REDD and UK-DALE datasets.  We follow a common established preprocessing procedure~\cite{zhang2018sequence}, which involves computing aggregate power from mains, aligning timestamps, and resampling all series to a uniform 6s interval via mean aggregation. Specifically, for the REDD dataset, the aggregate power is the sum of two separate mains channels, whereas for UK-DALE, a single mains channel is used. Following resampling, any resulting gaps in the time series are filled using a backfill method with a one-step limit to ensure data continuity. For experiments involving deep learning models, both the aggregate and appliance power data are standardized using pre-calculated mean and standard deviation values. Ground truth appliance states (ON=1, OFF=0) are generated using fixed power thresholds: microwave (200W), fridge (50W), dishwasher (10W), washing machine (20W), and kettle (2000W, only used in the UK-DALE dataset).

\begin{figure*}[htbp]
    \centering 
    \begin{subfigure}[b]{0.24\textwidth}
        \includegraphics[width=\textwidth]{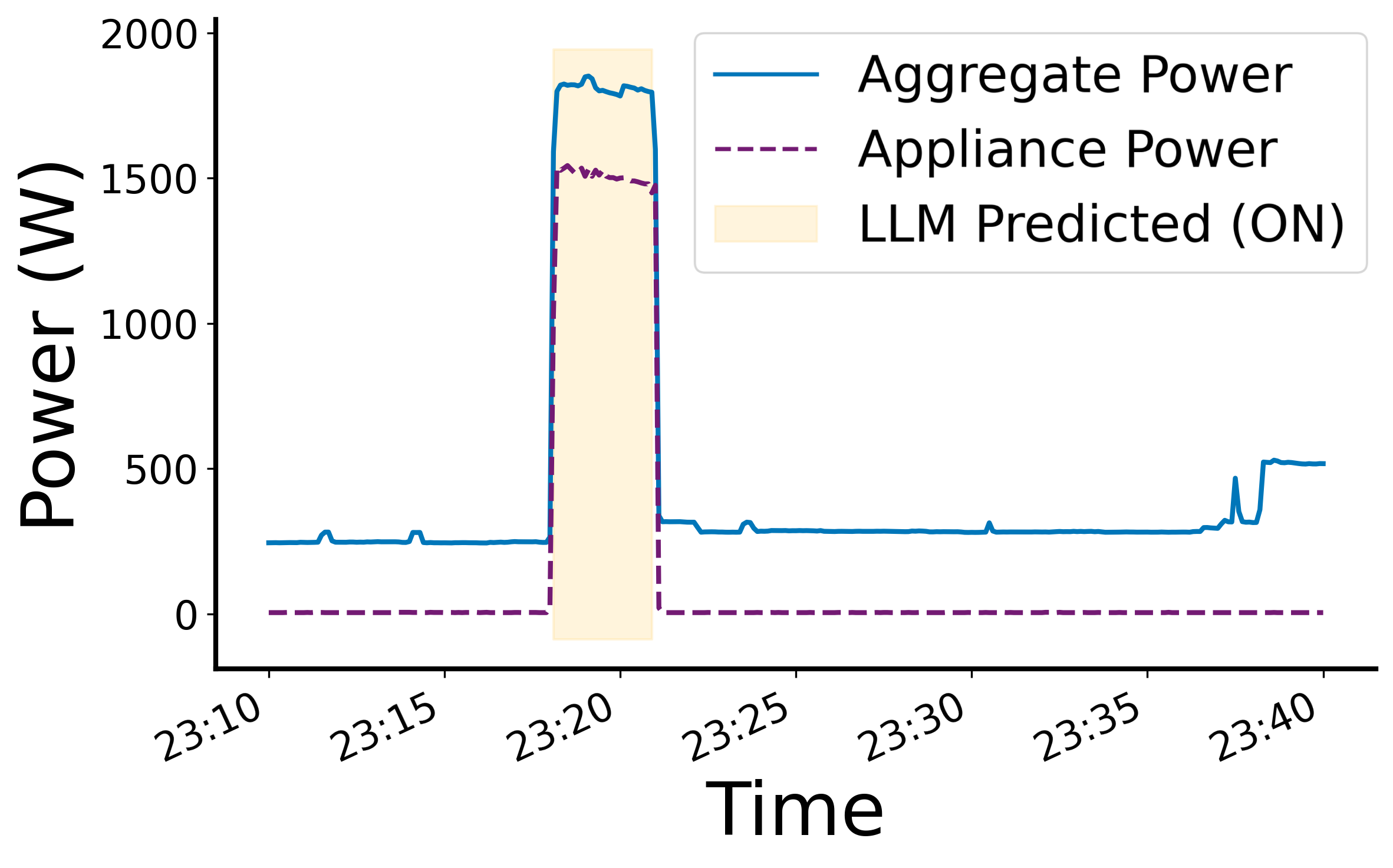}
        \caption{Microwave}
        \label{fig:microwave} 
    \end{subfigure}
    \hfill 
    \begin{subfigure}[b]{0.24\textwidth}
        \includegraphics[width=\textwidth]{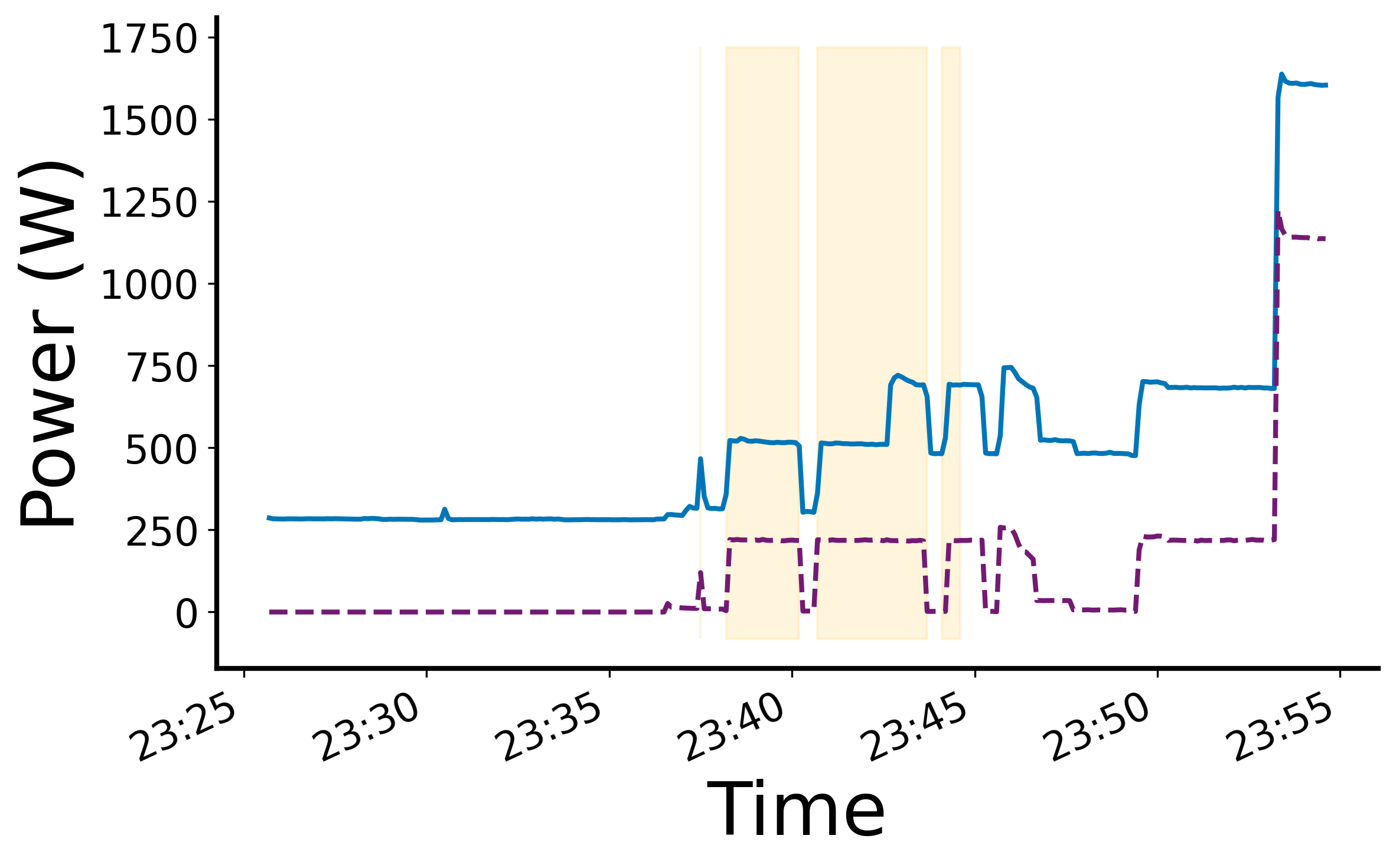}
        \caption{Dishwasher}
        \label{fig:dishwasher}
    \end{subfigure}
    \hfill 
    \begin{subfigure}[b]{0.24\textwidth}
        \includegraphics[width=\textwidth]{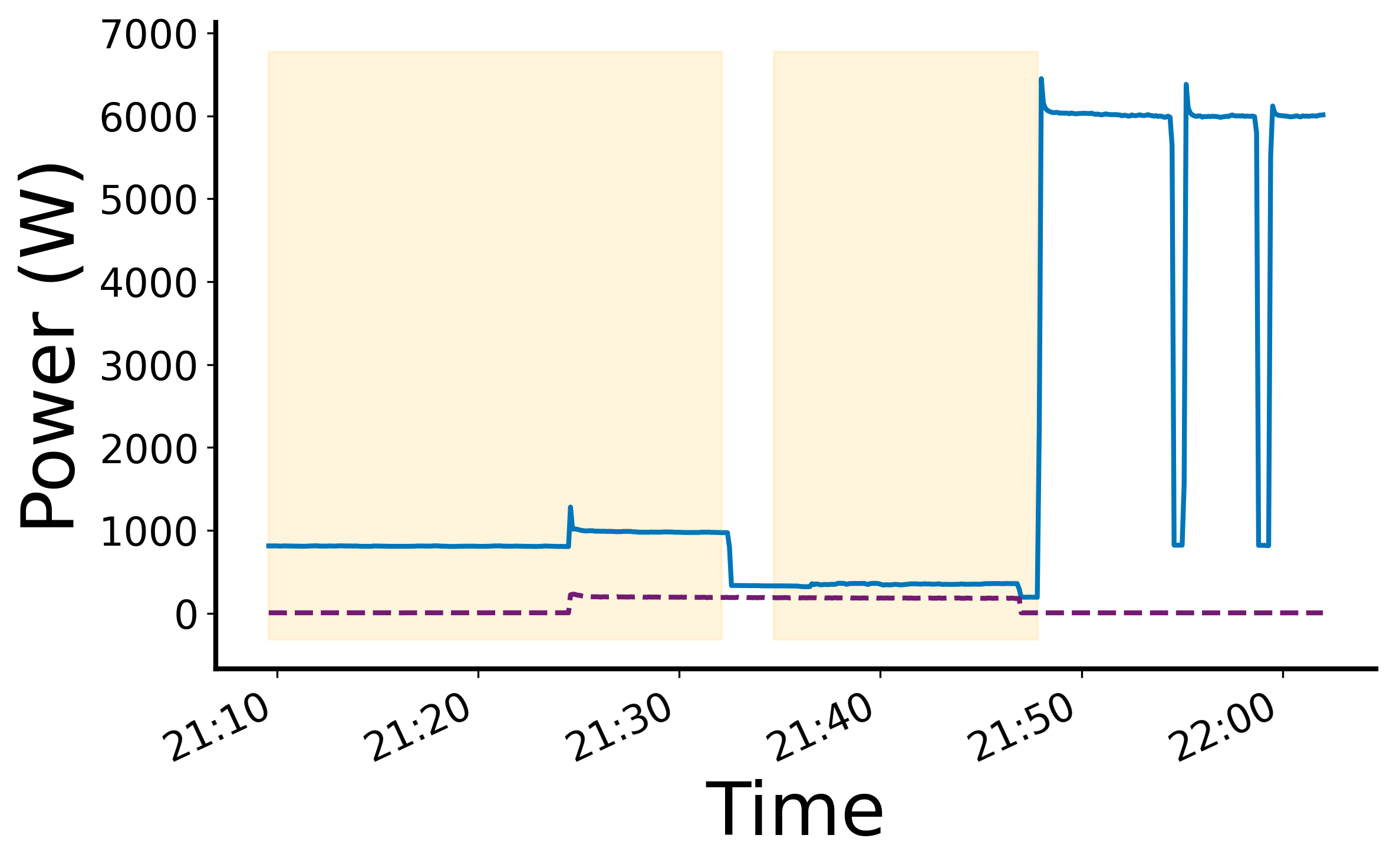}
        \caption{Fridge}
        \label{fig:fridge}
    \end{subfigure}
    \hfill 
    \begin{subfigure}[b]{0.24\textwidth}
        \includegraphics[width=\textwidth]{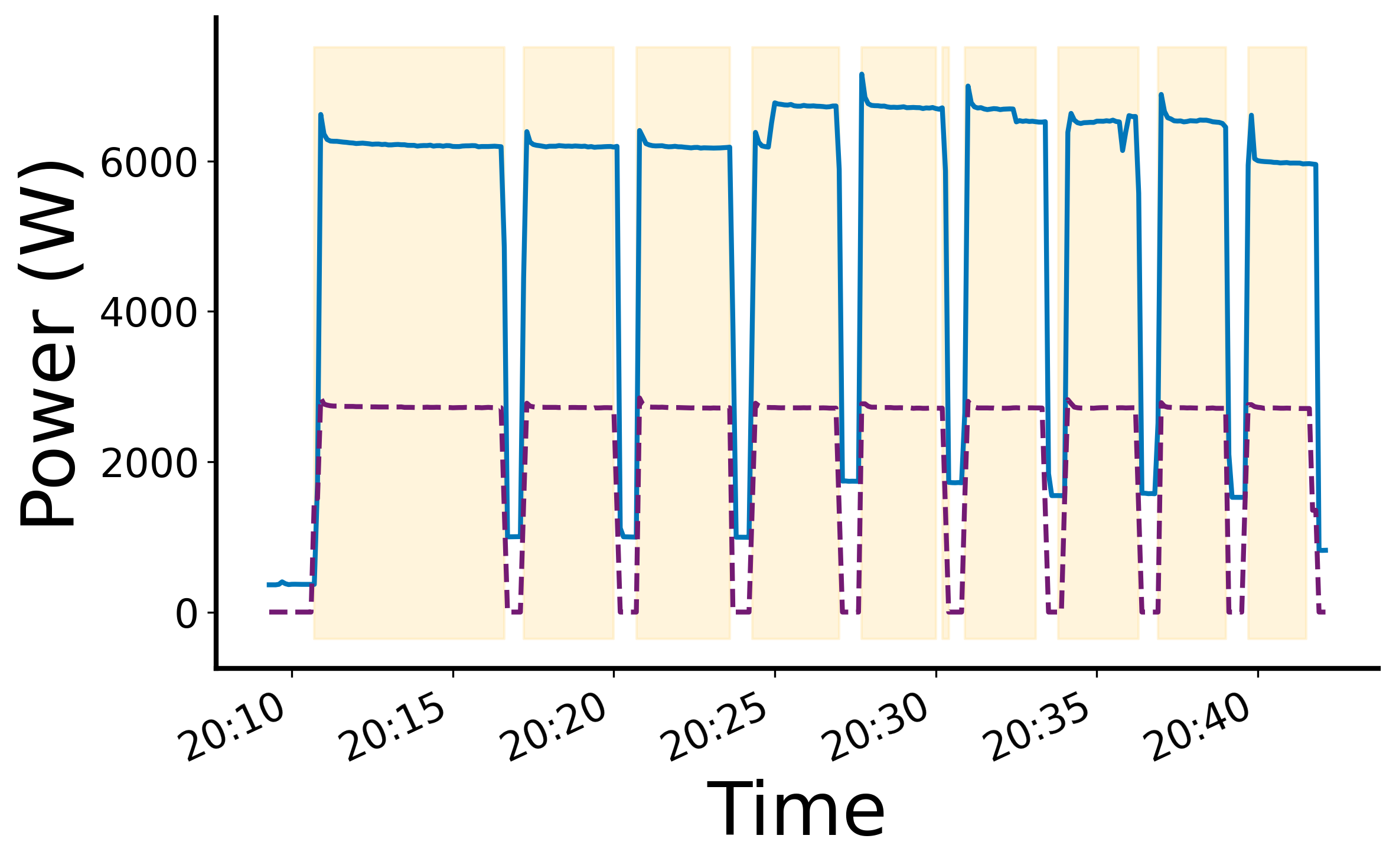}
        \caption{Washing Machine}
        \label{fig:washingmachine}
    \end{subfigure}
    \caption{Illustrative predictions on a one-day segment of the REDD house 1 test set. The results are generated by DeepSeek-V3-0324 using the optimized prompt.} 
    \label{fig:redd-vis-4-appliances}
    \vspace{-0.2cm}
\end{figure*}

\begin{figure*}[htbp]
    \centering 
    \begin{subfigure}[b]{0.24\textwidth}
        \includegraphics[width=\textwidth]{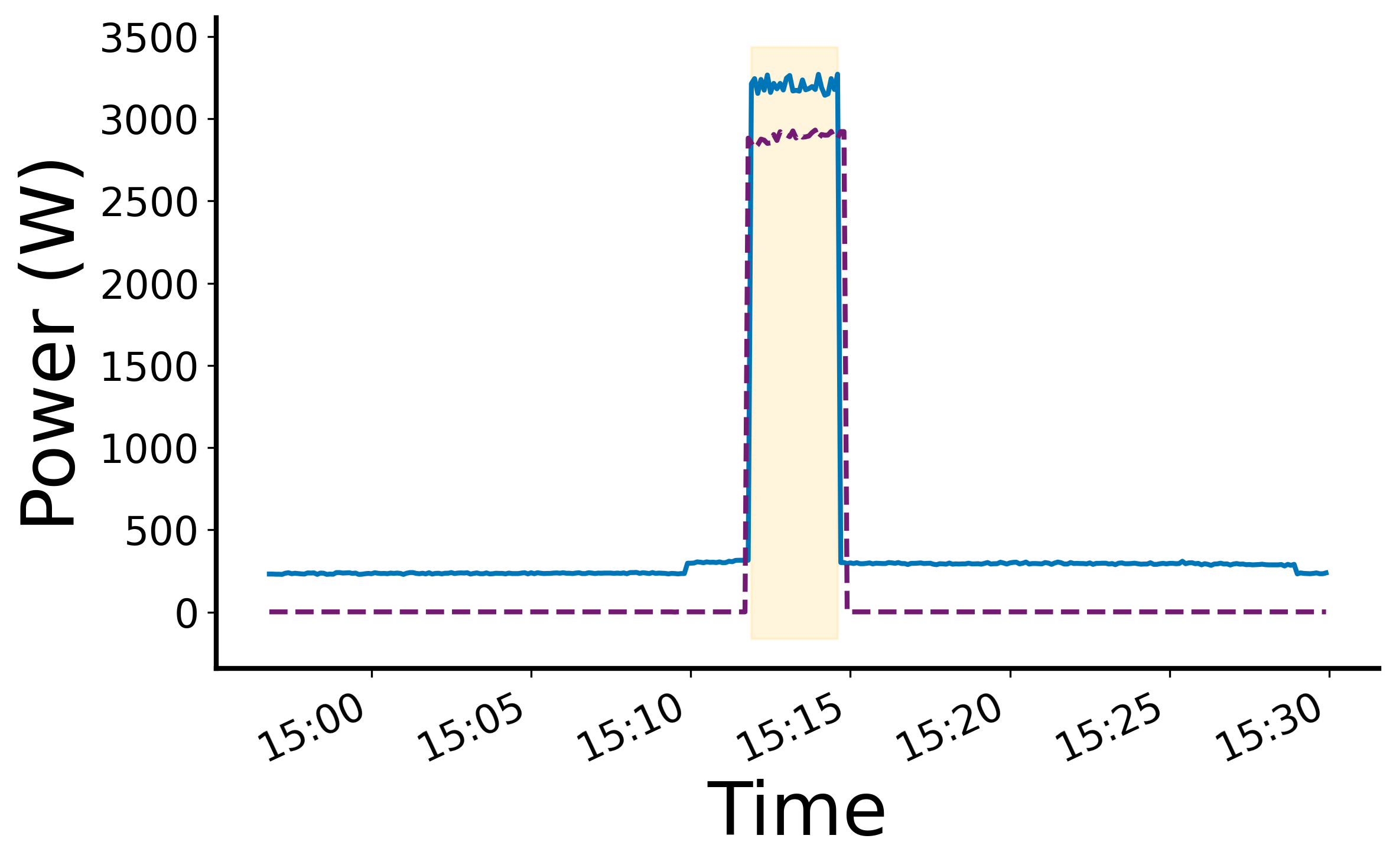}
        \caption{Kettle}
        \label{fig:microwave} 
    \end{subfigure}
    \hfill 
    \begin{subfigure}[b]{0.24\textwidth}
        \includegraphics[width=\textwidth]{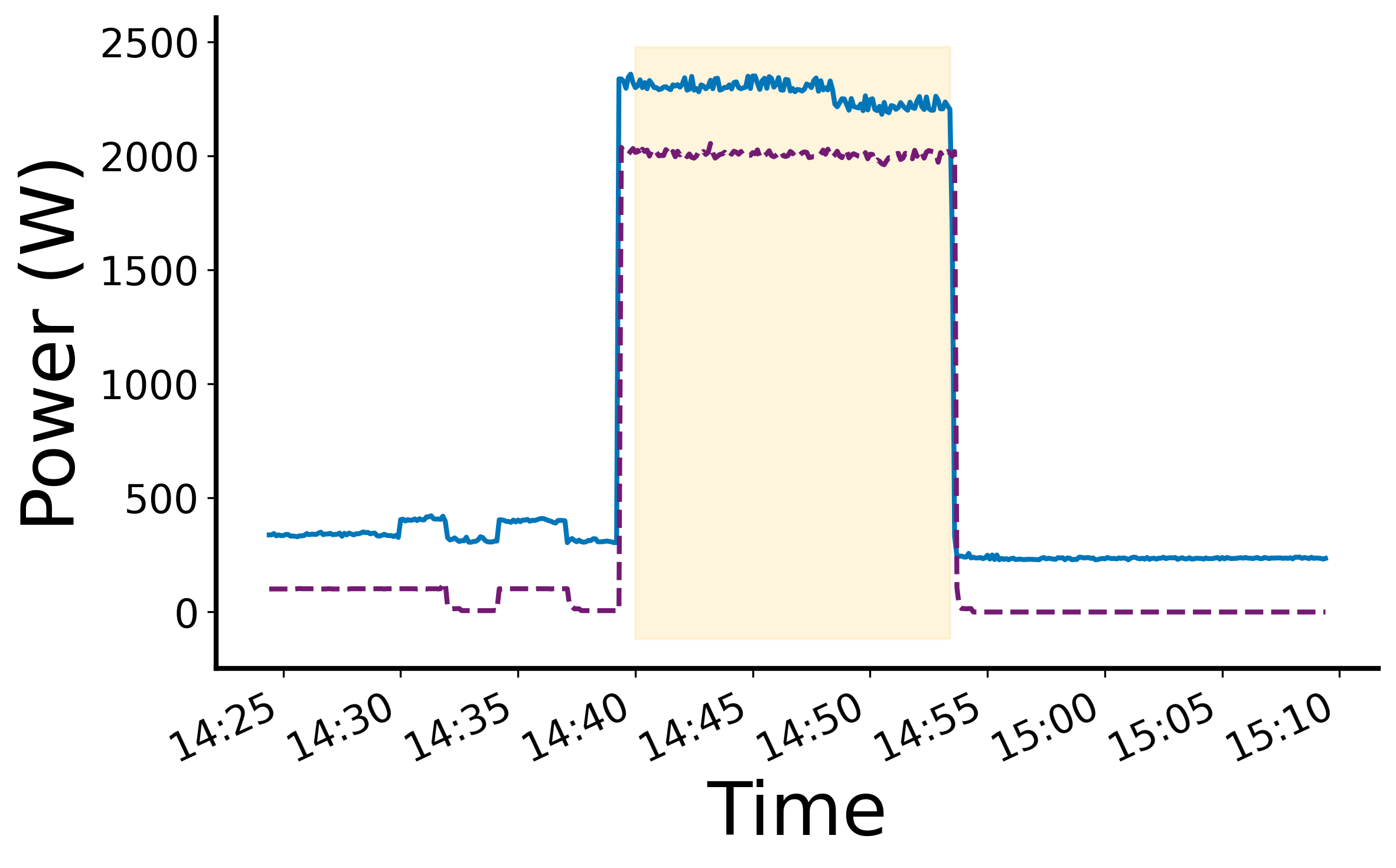}
        \caption{Dishwasher}
        \label{fig:dishwasher}
    \end{subfigure}
    \hfill 
    \begin{subfigure}[b]{0.24\textwidth}
        \includegraphics[width=\textwidth]{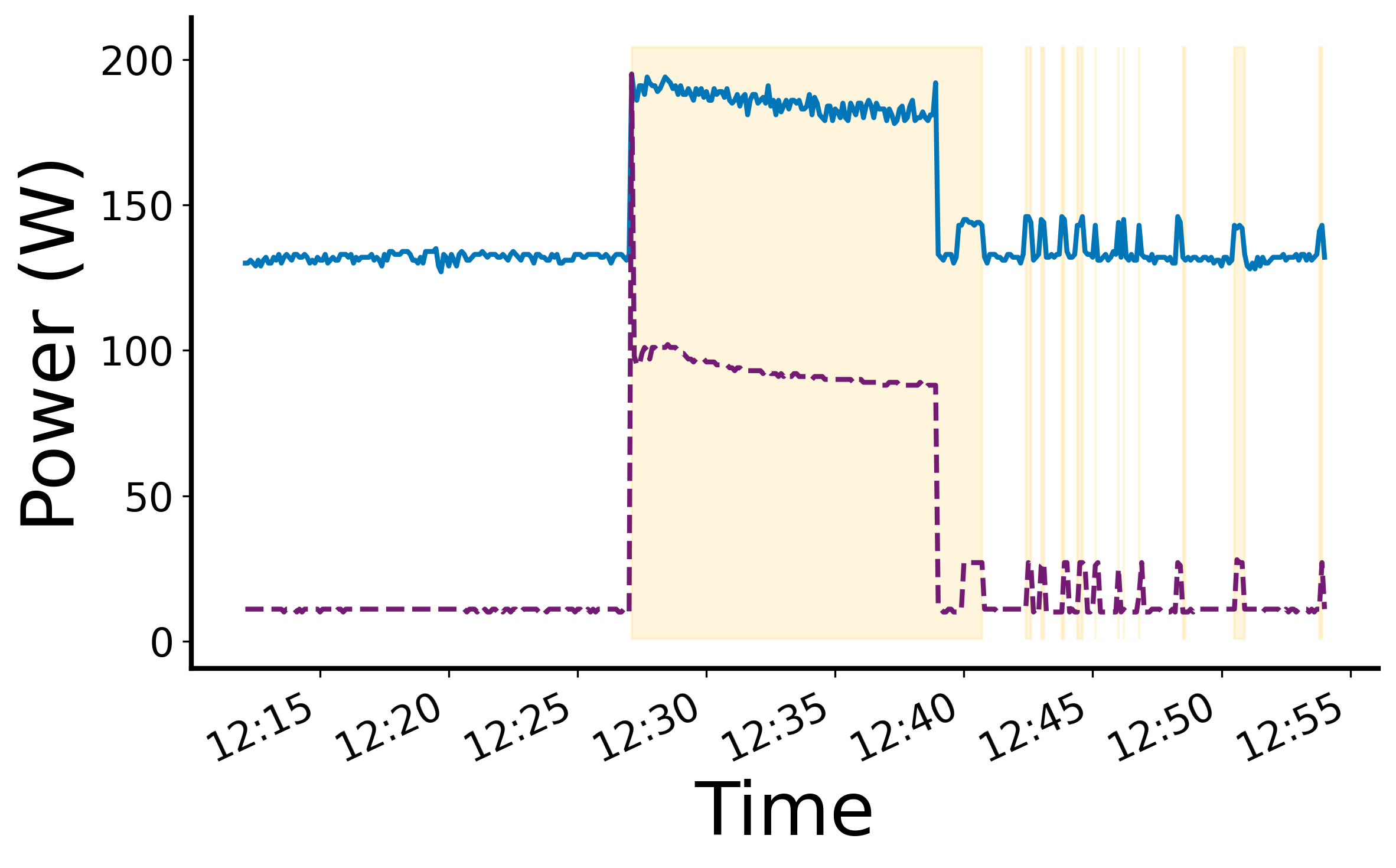}
        \caption{Fridge}
        \label{fig:fridge}
    \end{subfigure}
    \hfill 
    \begin{subfigure}[b]{0.24\textwidth}
        \includegraphics[width=\textwidth]{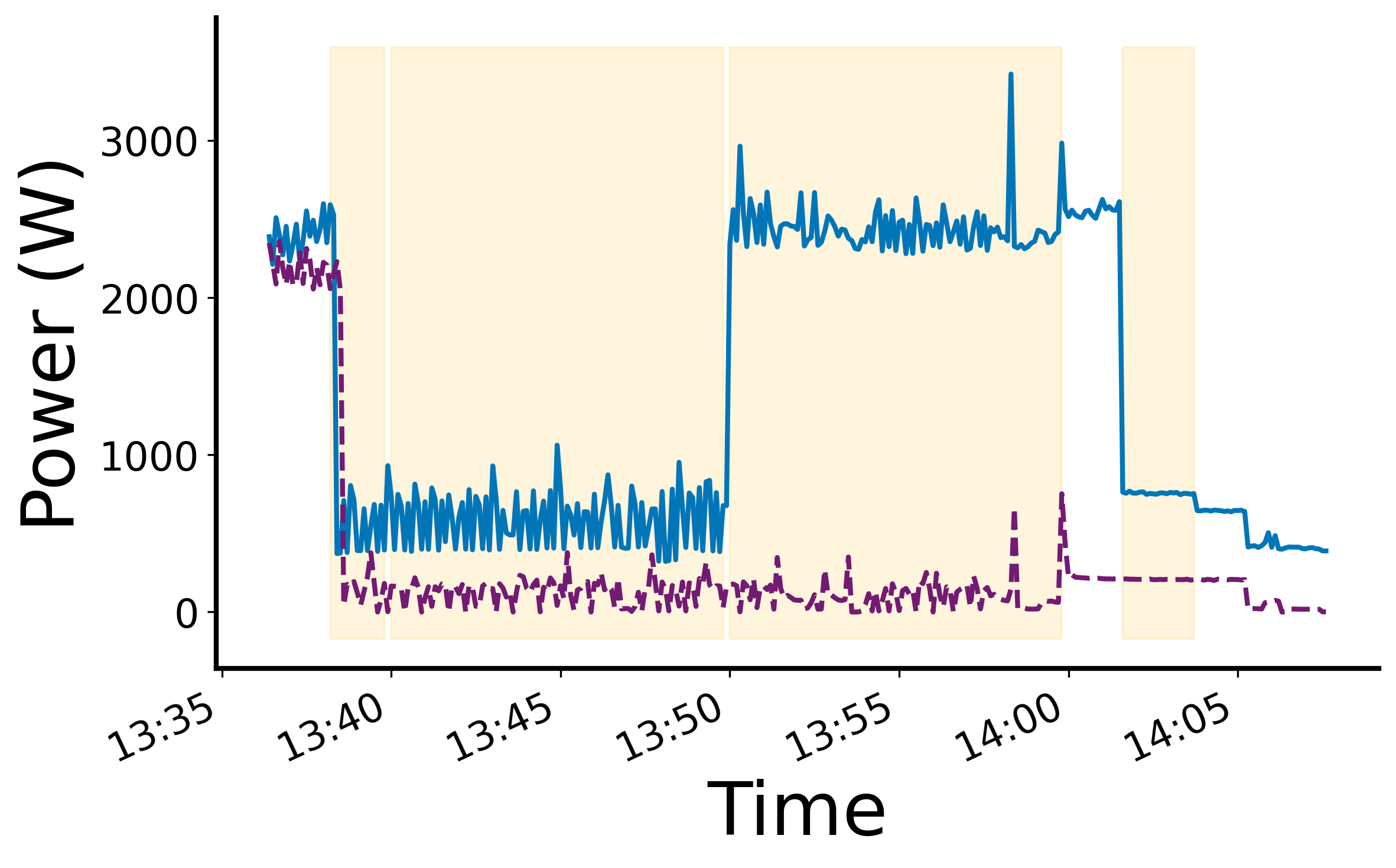}
        \caption{Washing Machine}
        \label{fig:washingmachine}
    \end{subfigure}
    \caption{Illustrative predictions on a one-day segment of the UK-DALE house 2 test set. The results are generated by DeepSeek-V3-0324 using the optimized prompt.} 
    \label{fig:ukdale-vis-4-appliances}
\end{figure*}

\subsection{Metrics}\label{subsec:metrics}
The performance of the appliance state detection task is evaluated using standard classification metrics. The evaluation is based on the counts of True Positives (TP), True Negatives (TN), False Positives (FP), and False Negatives (FN) calculated over the evaluation period.

While accuracy is intuitive, it can be misleading in NILM tasks. This is due to class imbalance, where the OFF state significantly dominates the ON state. Therefore, metrics that focus on the detection of the minority (ON) state are often more informative.

To provide a single, balanced assessment, especially when dealing with imbalanced datasets like those in appliance state detection, the \textbf{F1-score} is commonly used. It is the harmonic mean of Precision and Recall, giving a better measure of the incorrectly classified cases than accuracy does. The F1-score is particularly useful for evaluating the performance concerning the minority ON state.
\begin{equation}
    \text{Precision} = \frac{\text{TP}}{\text{TP} + \text{FP}}
\end{equation}

\begin{equation}
    \text{Recall} = \frac{\text{TP}}{\text{TP} + \text{FN}}
\end{equation}

\begin{equation}
    \text{F1-score} = \frac{2 \times \text{Precision} \times \text{Recall}}{\text{Precision} + \text{Recall}} = \frac{2 \times \text{TP}}{2 \times \text{TP} + \text{FP} + \text{FN}}
\end{equation}

\subsection{\label{subsec:modelselect}Model Selection, Task Focus, and Output Normalization}
Currently, mainstream large models are broadly categorized into two types: general models, such as GPT-4.1 and DeepSeek-V3, and reasoning models, like GPT-o3 and DeepSeek-R1~\cite{zhao2023survey}. Our pilot tests show that reasoning models, or general models prompted with chain-of-thought, tend to engage in extensive and unnecessary computational steps without yielding a corresponding improvement in performance. Consequently, this paper primarily evaluates general models, encompassing both dense and Mixture-of-Experts (MoE) architectures. The results of our tests on leading general models are presented in Figure~\ref{fig:sota-models}. For all subsequent case studies, we employed DeepSeek-V3-0324 for its superior performance and the lightweight GPT-4.1-mini for its efficiency.

Traditional NILM tasks typically encompass both state prediction and power estimation. However, our preliminary experiments indicate that achieving accurate power value predictions solely through prompting presents significant challenges. Specifically, on a single-day test slice of the REDD dataset, the DeepSeek-V3-0324 model records a mean SAE (Signal Aggregate Error, which measures the error in power estimation) of about 0.425, whereas the Seq2Point baseline attains a much lower SAE of 0.170. Given these quantitative results, we focus our subsequent experiments and primary analysis on the state detection task. This decision is also motivated by the fact that for practical NILM applications, identifying the operational state is the main objective, as average power consumption can often be inferred from the state itself.

For post-processing, we enable the JSON mode available in the inference API to ensure the returned results are parsable. Due to the inherent uncertainty in the outputs of large models, we also design an Output Normalizer module to improve the usability of the results. Specifically, two types of errors can occur in the LLMs' output. \romannumeral 1) Misaligned: the number of returned data points does not match the expected amount. We consider this error manageable and correct it using forward padding or truncation. \romannumeral 2) Malformed: the returned format is entirely incorrect and cannot be processed.

\section{Case Studies and Analysis}\label{sec:casestudies}
In this section, we give a detailed analysis of LLM4NILM to confirm the soundness of its design. We run ablation experiments (Section~\ref{subsec:ab-prompt}-\ref{subsec:ab-knowledge}) on every module and on the knowledge-injection process, and we conduct sensitivity studies on both window length and context length (Section~\ref{subsec:sensi-windowsize}-\ref{subsec:tradeoff}). Guided by the preliminary results in~\Cref{sec:implementation}, we choose DeepSeek-V3-0324 and GPT-4.1-mini as the representative models. For the case studies, we use one day of data from the REDD dataset, which contains about 2300 records during which four appliances are active.

\subsection{Ablation Study of Prompt Components}\label{subsec:ab-prompt}
According to~\Cref{sec:promptingLLMsforNILM}, we evaluate the effectiveness of each component of the foundational prompt. The window size and the context length are set to 100 and 10, respectively.

Our analysis begins with a foundational prompt (Base) containing only the Role \& Task and Input Data. 
Then we add the One-Shot Example and Knowledge Injection components to evaluate their impact on the NILM task.
Based on the foundational Prompt, we utilize timestamp and contextual information to enhance it.
For each configuration, we assess both the F1-score and the associated token cost.

Table~\ref{tab:prompts-results-nilm} presents the results for the DeepSeek-V3-0324 model. The data shows that adding the One-Shot Example, Knowledge Injection, and Context Information all enhance model performance. The improvements from knowledge injection and context information are the most significant. Counter-intuitively, combining timestamps and contextual information leads to a decrease in performance compared to using context alone. This suggests the LLM may encounter difficulty processing and prioritizing between absolute time and recent state history, potentially due to information overload.

Although adding components increases token usage, our findings indicate a clear optimal configuration. The prompt combining the Base, One-Shot Example, Knowledge Injection, and Contextual information (Base+OE+KI+CT) provides the most favorable performance-to-cost trade-off. 
Figure~\ref{fig:redd-vis-4-appliances} provides a visual demonstration of our optimized prompt's effectiveness, illustrating its predictions on a representative one-day segment from the unseen REDD test set. This visualization confirms the model's ability to identify key appliance states. Similarly, after adapting the prompt's knowledge for the UK-DALE dataset, Figure~\ref{fig:ukdale-vis-4-appliances} showcases comparable performance on a sample from its respective test set, further validating our prompt design.

\begin{tcolorbox}[insightbox]
\textbf{Insight 1.} {With carefully crafted prompt engineering and sufficiently descriptive features, LLMs demonstrate fundamental NILM capability, i.e., accurately detecting the ON/OFF transitions of individual appliances.}
\end{tcolorbox}

\begin{figure*}[htbp]
    \centering 
    \begin{subfigure}[b]{0.45\textwidth}
        \includegraphics[width=\textwidth]{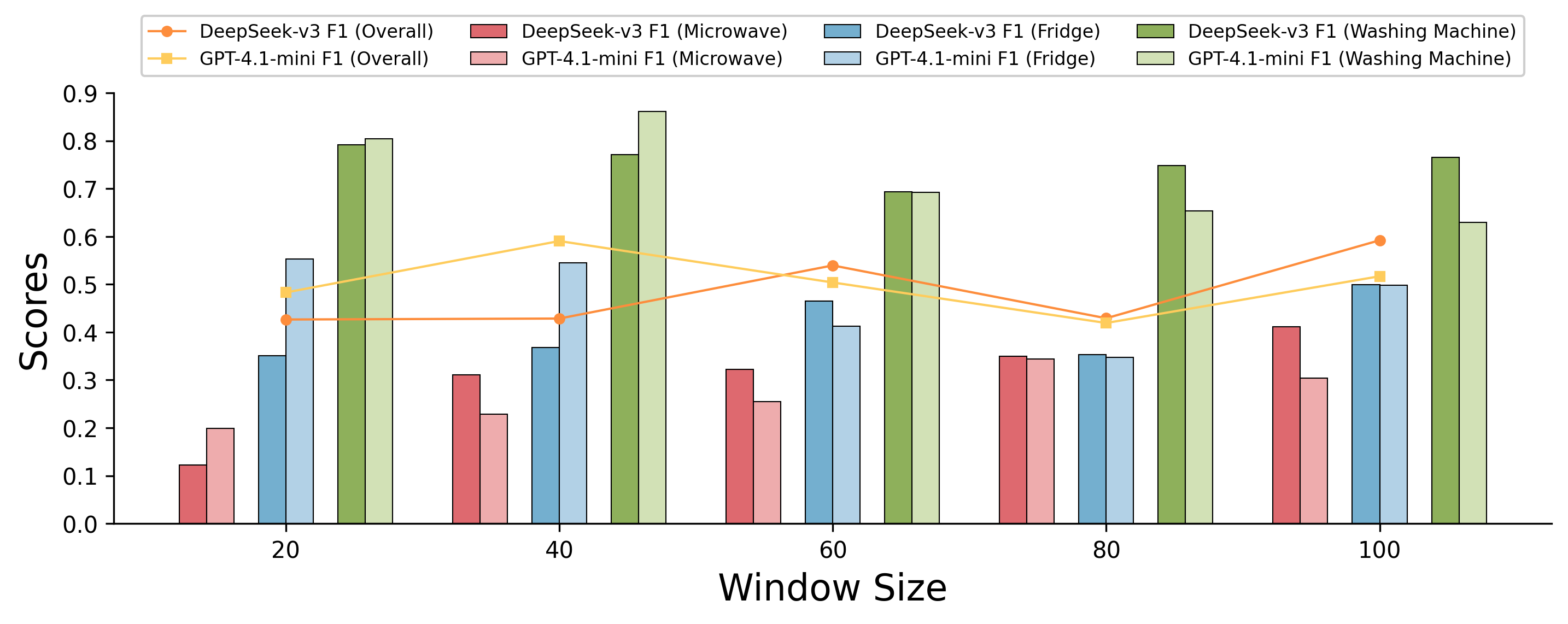}
        \caption{\label{fig:sensi-window-size}Impact of Window Size}
    \end{subfigure}
    \hfill 
    \begin{subfigure}[b]{0.45\textwidth}
        \includegraphics[width=\textwidth]{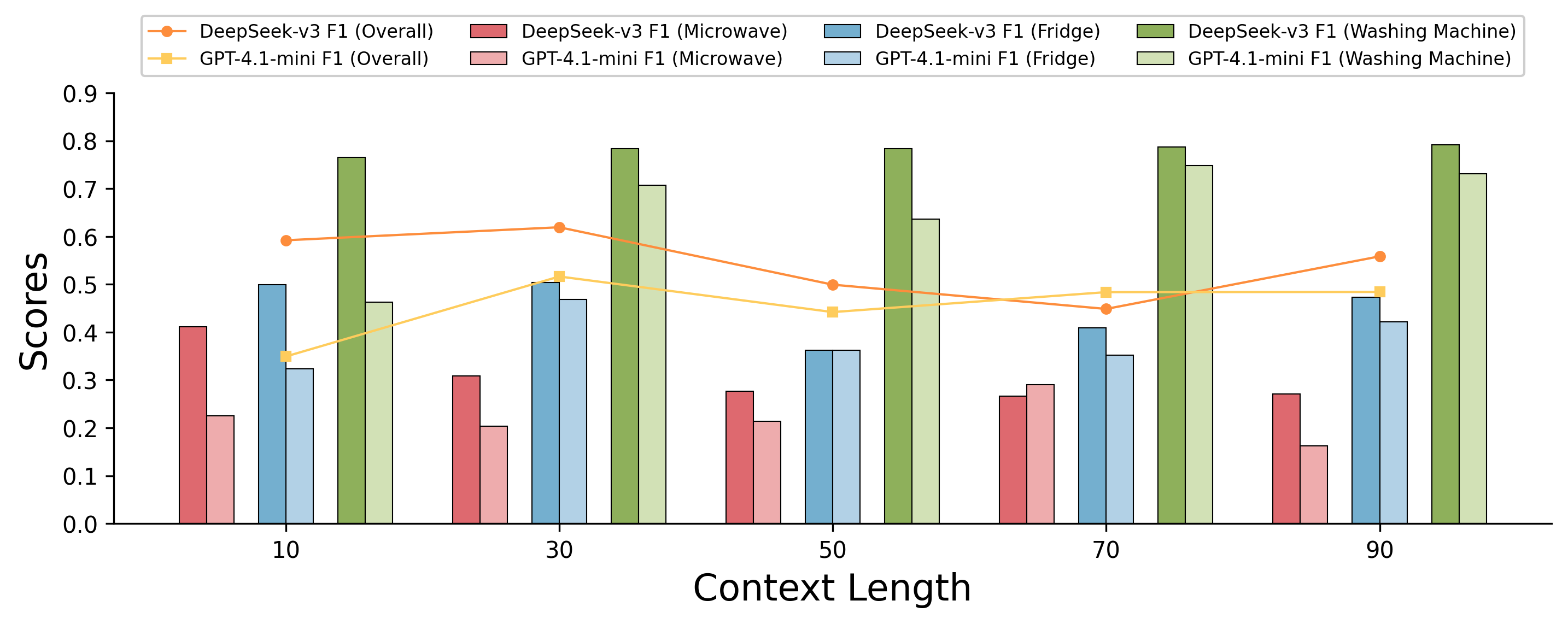}
        \caption{\label{fig:sensi-context}Impact of Context Length}
    \end{subfigure}
    \caption{Sensitivity analysis results for window size (left) and context length (right).} 
\end{figure*}

\subsection{Ablation Study of Knowledge Injection}\label{subsec:ab-knowledge}

\begin{table}[t]
  \centering
  \setlength{\tabcolsep}{3pt}               
  \caption{Overall impact of different knowledge–injection combinations on NILM state detection. \(\uparrow\): higher is better.}
  \label{tab:ablation-six}

  \begin{tabular*}{\linewidth}{@{\extracolsep{\fill}} ccc|ccc}
    \toprule
    \textbf{Power} & \textbf{Duration} & \textbf{Pattern} &
    \textbf{Precision} \(\uparrow\) & \textbf{Recall} \(\uparrow\) & \textbf{F1-score} \(\uparrow\)\\
    \midrule
    ✗ & ✓ & ✓ & 0.2548 & 0.3670 & 0.3008 \\
    ✓ & ✗ & ✓ & 0.4961 & 0.5801 & 0.5348 \\
    ✓ & ✓ & ✗ & 0.4656 & 0.7434 & 0.5725 \\
    ✓ & ✗ & ✗ & 0.4813 & 0.7097 & 0.5376 \\
    ✗ & ✓ & ✗ & 0.2896 & \textbf{0.7784} & 0.4222 \\
    ✗ & ✗ & ✓ & 0.3539 & 0.5597 & 0.4336 \\
    ✓ & ✓ & ✓ & \textbf{0.5246} & 0.6795 & \textbf{0.5921} \\
    \bottomrule
  \end{tabular*}
  \vspace{-0.5cm}
\end{table}

Domain knowledge injected into the prompt can be classified into three categories: power range, typical duration, and pattern descriptions. The result is presented in Table~\ref{tab:ablation-six}.

\textbf{Power Range.} When the prompt includes the power range, the model achieves an F1-score of 0.5376 even in the absence of any other information. In contrast, removing this cue reduces the F1 score to 0.3008. These results indicate that the power range is the principal factor that enables the LLM to differentiate appliance states.

\textbf{Pattern Descriptions.} Removing the pattern description from the complete prompt lowers the F1-score to 0.5725. When the model receives only the pattern description, the F1-score further drops to 0.4336. These findings show that the pattern description strengthens the model’s ability to identify appliances that share similar power levels but differ in burstiness or periodic behavior.

\textbf{Typical Duration.} The duration component offers the smallest gain when used in isolation. However, removing it from the full prompt reduces the F1-score to 0.5348. Its main role is to help the model distinguish appliances with long operating times.

\begin{tcolorbox}[insightbox]
\textbf{Insight 2.} {Power range is the most critical source of domain knowledge. Pattern and duration work as complementary factors. The combination of all three components allows the model to capture both quantitative and qualitative appliance features, which leads to the highest overall performance.}
\end{tcolorbox}

\subsection{Sensitivity Analysis of Window Size}\label{subsec:sensi-windowsize}
Window size refers to the length of the aggregate power sequence sent to the LLMs during a single inference pass. The LLMs then output, for each appliance, a state sequence of equal length.
We assess the influence of window lengths ranging from 20 to 100 on NILM accuracy.
Figure~\ref{fig:sensi-window-size} presents the F1-score for each appliance as well as the overall score.
Because the dishwasher is recorded only rarely, some models do not identify any of its activations at specific window lengths. We therefore omit the dishwasher from the figure, although the overall F1 still includes it.

DeepSeek-V3-0324 shows a clear upward trend. Its overall F1 rises from about 0.4265 at a window of 20 to a peak of 0.5921 at 100. GPT-4.1-mini behaves differently. It reaches its best F1 of 0.5905 at a window of 40, then falls steadily to 0.5169 at 100.
These results indicate that a longer window does not guarantee higher accuracy.
A moderate window of about forty samples balances context coverage and token budget for the lighter GPT-4.1-mini, whereas the larger DeepSeek-V3-0324 continues to benefit from added context up to one hundred samples before saturation appears.
Selecting window length should therefore match model capacity rather than copying the wide-window practice of conventional sequence models.

\begin{tcolorbox}[insightbox]
\textbf{Insight 3.} {Contrary to the intuition that a larger window always yields more information and thus higher accuracy, our results show that LLMs of different parameter scales exploit the information within a single window to varying degrees, so the optimal window length must be selected in line with the model size.}
\end{tcolorbox}

\subsection{Sensitivity Analysis of Context Length}\label{subsec:sensi-context}
Context length is the number of appliance‐state time steps from the preceding inference round that are fed into the model in the current round. 
Figure~\ref{fig:sensi-context} illustrates how the length of historical context affects NILM performance. In this analysis, we vary the context length from 10 to 90 while keeping the prompt structure consistent. 

Contrary to the intuition that more context improves accuracy, our results reveal a non-monotonic relationship for both models. DeepSeek-V3-0324 reaches its peak overall F1-score of 0.6193 at a context length of 30, while GPT-4.1-mini obtains its highest score of around 0.5164 at the same point. Providing additional context beyond this optimal length does not yield significant improvements and can even degrade performance.

\begin{tcolorbox}[insightbox]
\textbf{Insight 4.} {An excessively long context enlarges the receptive field but also introduces noise and dilutes the LLMs' focus instead of delivering purely useful signals. A moderate context length therefore strikes the best balance between information value and computational efficiency.}
\end{tcolorbox}

\subsection{Balancing Window Size and Model Capacity}\label{subsec:tradeoff}
Section~\ref{subsec:sensi-windowsize} has analyzed the influence of window size on NILM accuracy. Besides affecting accuracy, a longer window also challenges the model’s ability to respect the required output schema, because both the prompt and the answer must now fit within the same context budget. An excessively long window prevents the LLMs from producing outputs with the required length, as detailed in Section~\ref{subsec:modelselect}. 

\begin{figure}[!t]
  \centering
  \begin{subfigure}[t]{0.47\linewidth} 
    \includegraphics[width=\linewidth]{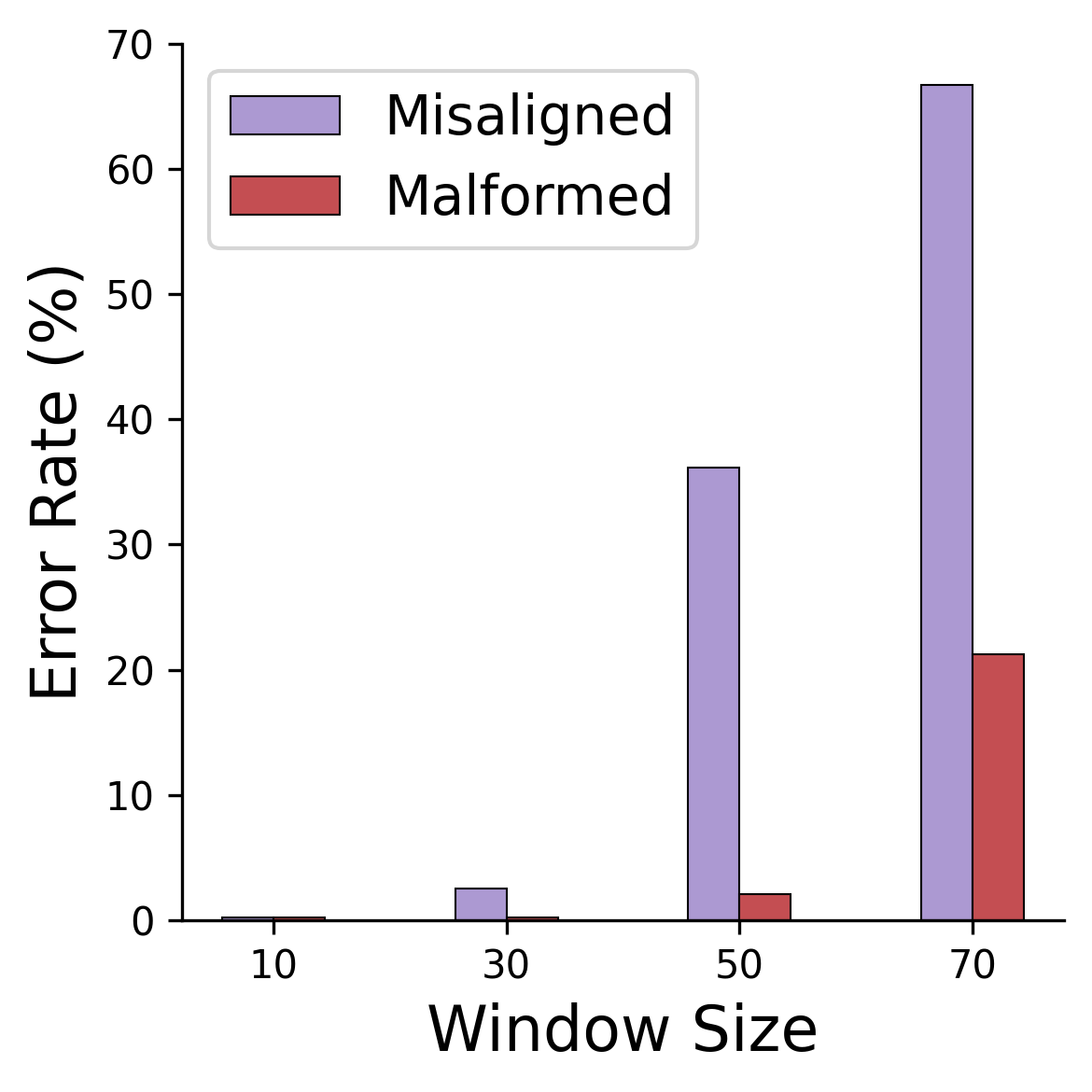}
    \caption{Llama-3.1-8B}
  \end{subfigure}
  \hfill 
  \begin{subfigure}[t]{0.47\linewidth}
    \includegraphics[width=\linewidth]{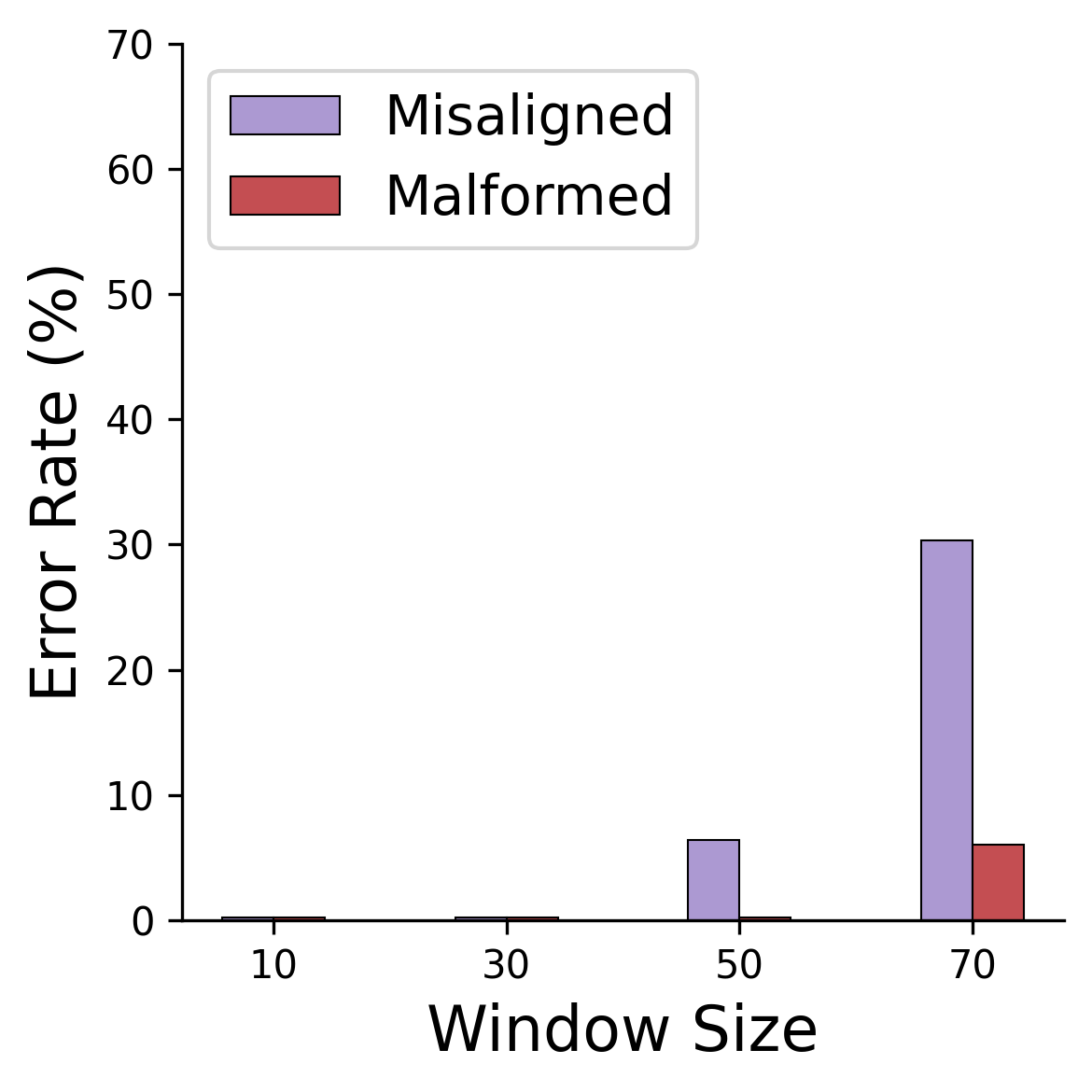}
    \caption{Llama-3.1-70B}
  \end{subfigure}
  \caption{Output Formatting Error Rates Across Different Model Scales.}
  \label{fig:error}

\end{figure}

Our earlier experiments with Deepseek-V3-0324 and GPT-4.1-mini revealed a stability gap for the same window size.
We therefore evaluate several Llama-3.1 models with different capacities. As shown in Figure~\ref{fig:error}, for the smaller Llama-3.1-8B model, both length errors and format errors increase sharply once the window length exceeds 30, with format errors being especially pronounced. In contrast, the larger Llama-3.1-70B model adapts better to large windows. Its error rate drops markedly, which shows clear robustness to format violations.

In short, higher capacity LLMs benefit from longer windows while still producing stable outputs, whereas lower capacity models need shorter windows to keep both accuracy and stability within acceptable limits.

\begin{tcolorbox}[insightbox]
\textbf{Insight 5.} {The choice of window length faces the dual requirement of preserving disaggregation accuracy and maintaining output stability. In real-world applications, practitioners need to select both the model capacity and the window length according to task demands so that accuracy and stability remain within acceptable bounds.}
\end{tcolorbox}

\section{Large-Scale Evaluation}
In \Cref{sec:casestudies}, we confirm the LLMs' basic capacity for NILM. Then, we proceed with a large-scale evaluation to assess its real-world usability (\Cref{subsec:effect}), generalization capabilities (\Cref{subsec:general}), and explainability (\Cref{subsec:explan}).

To ensure a rigorous evaluation, we establish the following experimental setup. For the REDD dataset, data from houses 2 and 3 are used to extract appliance knowledge for prompts and to train the DL baselines. The unseen house 1, containing 262,727 records, serves as the exclusive test set. For the UK-DALE dataset, we use data from house 2, splitting it into training (70\%), validation (10\%), and testing (20\%) sets, with the test set comprising 345,565 records. We use \textbf{Seq2Seq} and \textbf{Seq2Point} with \textbf{CNNs}~\cite{zhang2018sequence} as representative baseline models, as they are well-established and commonly benchmarked approaches in the NILM field. For the LLM-based approach, we use the optimized prompt strategy from~\Cref{sec:promptingLLMsforNILM} with a window size of 100 and a context length of 10.

\subsection{Effectiveness Analysis (\textbf{RQ1})}\label{subsec:effect}
Table~\ref{tab:redd_ukdale_compact} presents the comparative results, revealing a significant performance gap between the specialized DL models and the prompt-based LLM approach. On the unseen REDD House 1, Seq2Point and Seq2Seq achieve overall F1-scores of 0.7289 and 0.7285, substantially outperforming DeepSeek-V3-0324 (0.4397). This trend is confirmed on UK-DALE House 2, where the DL models again lead with F1-scores of 0.8721 and 0.8761, compared to 0.6120 from DeepSeek-V3-0324.

A deeper analysis reveals that LLMs excel at identifying appliances with highly distinctive power signatures. The washing machine in REDD and the kettle in UK-DALE, which have the highest peak power consumption in their respective datasets (approx. 6000W and 3000W), are among the most accurately identified appliances by the LLMs. These large, sharp power draws create unambiguous events in the aggregate signal that align well with the simplified descriptions in the prompts.

The fridge presents a unique case. Despite its moderate power draw, it is identified with consistent success across both datasets (F1-scores of 0.4802 on REDD and 0.6182 on UK-DALE). This success is not due to peak power but to its strong periodic cycling and long operational duration. The LLM learns to recognize this recurring pattern from the provided knowledge. Conversely, appliances with more complex signatures pose a challenge. The dishwasher, for example, operates in multiple stages with varying power levels. While its high-power cycles may be detected, its medium and low-power stages are often masked by other loads, leading to a lower F1-score.

The overall performance gap stems from the inherent limitations of a prompt-only approach. The prompts are data-free and encode only coarse appliance characteristics. The models struggle when an appliance's real-time operation deviates from these simplified descriptions. As general-purpose models, LLMs cannot fully internalize the complex physics of power signals from natural language alone. This explains the accuracy drop compared to specialized models trained directly on power data.

\begin{table*}[htbp]
  \centering
  \caption{Model performance on REDD house 1 and UK‑DALE house 2. MW, FR, WM, DW, and KT denote the microwave, fridge, washing machine, dishwasher, and kettle, respectively. $\uparrow$: higher is better.}
  \label{tab:redd_ukdale_compact}
  \scriptsize
  \resizebox{0.95\textwidth}{!}{%
  \begin{tabular}{l l | *{5}{c}| *{6}{c}}
    \toprule
    \multirow{2}{*}{\textbf{Model}} & \multirow{2}{*}{\textbf{Metric}}
      & \multicolumn{5}{c|}{\textbf{REDD}} & \multicolumn{6}{c}{\textbf{UK‑DALE}} \\
    \cmidrule(lr){3-7}\cmidrule(lr){8-13}
    & & MW & FR & WM & DW & Overall & MW & FR & WM & DW & KT & Overall \\
    \midrule
    \multirow{3}{*}{Seq2Seq}
      & Precision $\uparrow$ & 0.4265 & 0.6263 & 0.5185 & 0.5646 & 0.5997 & 0.7766 & 0.8641 & 0.3371 & 0.5953 & 0.9083 & 0.8071 \\
      & Recall $\uparrow$    & 0.6306 & 0.9712 & 0.9981 & 0.8200 & 0.9277 & 0.9906 & 0.9671 & 0.9669 & 0.9400 & 0.9973 & 0.9656 \\
      & F1‑score $\uparrow$  & 0.5088 & 0.7615 & 0.6825 & 0.6687 & 0.7285 & 0.8706 & 0.9127 & 0.4999 & 0.7289 & 0.9507 & 0.8761 \\
    \midrule
    \multirow{3}{*}{Seq2Point}
      & Precision $\uparrow$ & 0.4531 & 0.6686 & 0.4573 & 0.5012 & 0.6168 & 0.7288 & 0.8246 & 0.4559 & 0.7587 & 0.9202 & 0.8026 \\
      & Recall $\uparrow$    & 0.8642 & 0.9277 & 0.9984 & 0.6421 & 0.8906 & 0.9906 & 0.9596 & 0.9265 & 0.8990 & 0.9973 & 0.9548 \\
      & F1‑score $\uparrow$  & 0.5945 & 0.7771 & 0.6273 & 0.5629 & 0.7289 & 0.8398 & 0.8870 & 0.6111 & 0.8229 & 0.9572 & 0.8721 \\
    \midrule
    \multirow{3}{*}{GPT-4.1-mini}
      & Precision $\uparrow$ & 0.1975 & 0.2461 & 0.2842 & 0.0091 & 0.2205
                   & 0.0794 & 0.3476 & 0.0470 & 0.0515 & 0.3546 & 0.3016 \\
      & Recall $\uparrow$    & 0.5131 & 0.8811 & 0.8139 & 0.0263 & 0.7554
                   & 0.5565 & 0.9974 & 0.2842 & 0.1573 & 0.8083 & 0.9047 \\
      & F1-score $\uparrow$  & 0.2852 & 0.3847 & 0.4213 & 0.0135 & 0.3414
                   & 0.1390 & 0.5155 & 0.0807 & 0.0776 & 0.4930 & 0.4524 \\
    \midrule
    \multirow{3}{*}{DeepSeek‑V3-0324}
      & Precision $\uparrow$ & 0.2035 & 0.3343 & 0.3594 & 0.1388 & 0.3093
                   & 0.5192 & 0.4683 & 0.3239 & 0.6802 & 0.4924 & 0.4739 \\
      & Recall $\uparrow$    & 0.3420 & 0.8525 & 0.7739 & 0.3060 & 0.7599
                   & 0.3996 & 0.9094 & 0.4633 & 0.5427 & 0.9078 & 0.8640 \\
      & F1-score $\uparrow$  & 0.2552 & 0.4802 & 0.4908 & 0.1909 & 0.4397
                   & 0.4516 & 0.6182 & 0.3812 & 0.6038 & 0.6385 & 0.6120 \\
    \bottomrule
  \end{tabular}}
  \normalsize
\end{table*}

\textbf{\textit{Answer to RQ1}. On NILM tasks, applying prompt engineering alone to current general LLMs still lags behind specialized deep learning models in large-scale tests. Nevertheless, these models support rapid zero-data deployment, and their accuracy can rise with targeted knowledge injection and adaptive prompting.}

\subsection{Generalization Analysis (RQ2)}\label{subsec:general}
Generalizability denotes a model’s capacity to retain reliable disaggregation accuracy when deployed in previously unseen households, geographical regions, or appliance types~\cite{d2019transfer}. 
Accordingly, we test the LLMs across new households, regions and appliance sets without any task-specific fine-tuning.

\textbf{Cross Houses.} 
For cross-household generalization, we apply the LLM to REDD house 1, an entirely unseen environment. The model is guided only by prompts containing appliance knowledge derived from houses 2 and 3. The results in Table~\ref{tab:redd_ukdale_compact} and Figure~\ref{fig:redd-vis-4-appliances} confirm that the model can identify the appliance states, demonstrating its ability to generalize within the same region.

\textbf{Cross Regions.} 
We extend the analysis to cross-region generalization using the UK-DALE dataset. The framework's adaptability is highlighted by its ability to handle this new region and its different appliance set (e.g., the kettle) solely by modifying the prompt's knowledge injection component. As shown in Table~\ref{tab:redd_ukdale_compact} and Figure~\ref{fig:ukdale-vis-4-appliances}, the LLM achieves effective performance without architectural changes or retraining. This plug-and-play capability stands in sharp contrast to conventional DL models, which require extensive retraining for new environments.

\textbf{\textit{Answer to RQ2}. Through appropriate prompting strategies, LLMs show a capacity for robust generalization in NILM across diverse and previously unseen environments. This capability offers an advantage where traditional DL models might otherwise require retraining or complex transfer learning techniques when deployed in new settings.}

\subsection{Explainability Analysis (RQ3)}\label{subsec:explan}
In NILM research, a model is considered explainable when a human can understand the reasons for each disaggregation decision~\cite{mollel2023explainability}.We examine whether LLMs can enhance the explainability of the NILM process by providing several justifications for each appliance.

Our methodology requires prompting the LLM to perform two actions on representative time-series snippets. First, it performs the disaggregation task. Second, it generates a textual explanation for the ON/OFF status it assigns to each appliance, as shown in Appendix~\ref{appendix:explanations}. We then evaluate these explanations by comparing them against the ground truth data and the aggregate power signal.

We find that LLMs generate plausible explanations in most cases. The explanations correctly identify specific patterns, such as sudden power spikes in the aggregate signal. They also correlate the data with known appliance characteristics provided in the prompt, like typical power draw or operational cycles. Although LLMs are themselves complex models, their ability to produce human-readable text provides a clear rationale for their outputs.

These explanations offer valuable insight into the model's decision-making process. Furthermore, even when an output is incorrect, the accompanying text can reveal the LLM's misinterpretations. This provides actionable feedback for refining the prompts to improve future performance.

\textbf{\textit{Answer to RQ3}. The inherent natural language processing capabilities of LLMs can contribute to enhancing NILM explainability by generating verifiable explanations for energy disaggregation outputs. This offers a step towards greater transparency in the decision-making process.}

\section{Discussion}\label{sec:discussion}
Our findings show that applying LLMs to NILM tasks via direct prompting yields limited accuracy. This result is anticipated, as the core architecture of an LLM is optimized for language, not for time-series signal decomposition, and prompt engineering alone appears to have reached its performance ceiling for this task. To enhance core NILM performance, future work could explore methods like fine-tuning on domain-specific datasets. This, however, introduces a clear trade-off, sacrificing the zero-data, training-free advantage for higher accuracy and increased computational cost. The true value of the current approach, therefore, lies not in competing on precision but in its remarkable generalization and explainability.

The full potential of LLM-based NILM is unlocked when it is integrated as an intrinsic capability of a smart home agent. Exploring the boundaries of LLMs is critical as these agents become central to future smart homes. An agent with foundational appliance-level awareness, even if imperfect, can create powerful synergies by fusing real-time energy data with its existing contextual knowledge, such as user schedules and preferences. For instance, it could not only identify that a washing machine is running but also recommend operating it during off-peak hours or, more critically, flag it as an anomaly if the user is known to be away. This form of context-aware energy management represents a significant leap beyond traditional NILM and is a capability that isolated, specialized systems struggle to provide.

\section{Conclusion}\label{sec:conclusion}
This paper presents the first systematic investigation into applying prompt-based, general-purpose LLMs to core NILM tasks. Through comprehensive prompt engineering and evaluation on established benchmarks, our findings reveal a clear trade-off. While carefully crafted prompts provide LLMs with foundational NILM capabilities, their disaggregation accuracy is not yet competitive with specialized deep learning models. However, we demonstrate that this prompt-only approach facilitates robust zero-shot generalization across diverse environments and enhances explainability by generating human-readable justifications for its predictions. Our work underscores that the immediate value of LLMs in this domain lies not in achieving state-of-the-art accuracy, but in leveraging their unique generalization and explainability. This charts a promising new direction for future research in creating more adaptable, transparent, and context-aware energy management systems.

\bibliographystyle{unsrt}
\bibliography{mainfiles/references}

\appendix
\section{Prompts}\label{appendix:prompts}
\textbf{Role}. You are an expert system specializing in Non-intrusive Load Monitoring (NILM). Your job is to analyze a sequence of aggregate active power readings and determine the ON (1) or OFF (0) status for each of the following appliances at every time step: \textit{APPLIANCE\_LIST}.

\noindent \textbf{Task}. Given an input sequence of exactly \textit{WINDOW\_SIZE} aggregate power readings, use the prior knowledge and context below to infer the ON or OFF status for each appliance at each time step. Sampling cycle is 6 s.

\noindent \textbf{Output Format}.

\noindent - Respond ONLY with a single JSON object.

\noindent - Keys must be exactly: "\textit{APPLIANCE\_NAMES[0]\_status}", ..., \textit{APPLIANCE\_NAMES[n]\_status}".

\noindent - Each value is a list of exactly integers (0 or 1).

\noindent - Do NOT output any explanation, extra text, markdown, or code block, ONLY the JSON object.

\noindent - If uncertain, make the best guess, but do NOT use any value other than 0 or 1.

\noindent - The output list for each appliance MUST have exactly \textit{WINDOW\_
SIZE} elements.

\noindent \textbf{Prior Knowledge.}

\noindent\textit{APPLIANCE\_NAMES[0]}: 

\noindent Stand-by Power, Power Range, Duration, Usage Pattern

\noindent\textit{APPLIANCE\_NAMES[1]}: 

\noindent Stand-by Power, Power Range, Duration, Usage Pattern

\noindent...

\noindent \textit{APPLIANCE\_NAMES[n]}: 

\noindent Stand-by Power, Power Range, Duration, Usage Pattern

\noindent \textbf{Example Input}. 

\noindent Aggregate Power: [597.540, 597.397, 597.752, 597.462,
437.508, 169.120, 169.324, 169.130, 169.112, 169.404]

\noindent \textbf{Example Output}.

\noindent ```json

\noindent "\textit{APPLIANCE\_NAMES[0]\_status}": [0,0,0,0,0,0,0,0,0,0],

\noindent "\textit{APPLIANCE\_NAMES[1]\_status}": [1,1,1,1,1,0,0,0,0,0],

\noindent "\textit{APPLIANCE\_NAMES[2]\_status}": [0,0,0,0,0,0,0,0,0,0],

\noindent ...

\noindent "\textit{APPLIANCE\_NAMES[n]\_status}": [0,0,0,0,0,0,0,0,0,0]

\noindent ```

\noindent \textbf{Context.}

\noindent The states predicted in the previous \textit{CONTEXT\_LENGTH} steps: \textit{PREVIOUS\_WINDOW\_STATES} 

\section{Explanations}\label{appendix:explanations}

\noindent \textbf{Microwave (ON)}. "microwave\_explanation": "The microwave shows a characteristic power increase at timestep 58, aligning with its known operational wattage from prior knowledge, and turns off shortly after."

\noindent \textbf{Microwave (OFF)}. "microwave\_explanation": "No significant power changes matching the microwave's profile (e.g., high power bursts around 1500W) were observed; state remains OFF."

\noindent \textbf{Fridge (OFF)}. "fridge\_explanation": "No significant power changes matching the fridge's profile (e.g., cyclical compressor activity around 150W) were observed; state remains OFF."

\noindent \textbf{Dishwasher (OFF)}. "dishwasher\_explanation": "No significant power changes matching the dishwasher's profile (e.g., multi-stage power levels) were observed; state remains OFF."

\noindent \textbf{Washing Machine (ON)}. "washingmachine\_explanation": "The washing machine shows repeated abrupt changes between high power (around 6000W) and low power (around 800W), consistent with its known operational pattern. It turns OFF at timestep 10, 40, and 66, and turns ON again shortly after each OFF period."

\noindent \textbf{Washing Machine (OFF)}. "washingmachine\_explanation": "No significant power changes matching the washing machine's profile (e.g., high power bursts around 2600-2700W) were observed; state remains OFF."

\end{document}